\documentclass{article}

\usepackage[final]{neurips_2023}

\usepackage[utf8]{inputenc} 
\usepackage[T1]{fontenc}    
\usepackage{hyperref}       
\usepackage{url}            
\usepackage{booktabs}       
\usepackage{amsfonts}       
\usepackage{nicefrac}       
\usepackage{microtype}      
\usepackage{xcolor}         
\usepackage{amsmath}
\usepackage[english]{babel}
\usepackage{caption}
\usepackage{multirow}
\usepackage{multicol}
\usepackage{tabularx}
\usepackage{natbib}
\setcitestyle{authoryear,open={(},close={)}}

\title{Comprehensive Bench-marking of Entropy and Margin Based Scoring Metrics for Data Selection}

\usepackage{graphicx}
\graphicspath{ {./Figures/} }

\author{%
Anusha Sabbineni$^{1}$ \quad Nikhil Anand$^{1}$ \quad Maria Minakova$^2$ \\
$^1$Alexa AI (Amazon) \quad $^2$Work done at Amazon\\
\texttt{\{sabanu,nkhlanan\}@amazon.com}\\
\texttt{\{maria.s.minakova\}@gmail.com}\\
}

\begin{document}

\maketitle


\begin{abstract}
     While data selection methods have been studied extensively in active learning, data pruning, and data augmentation settings, there is little evidence for the efficacy of these methods in industry scale settings, particularly in low-resource languages. Our work presents ways of assessing prospective training examples in those settings for their "usefulness" or "difficulty". We also demonstrate how these measures can be used in selecting important examples for training supervised machine learning models. We primarily experiment with entropy and Error L2-Norm (EL2N) scores. We use these metrics to curate high quality datasets from a large pool of \textit{Weak Signal Labeled} data, which assigns no-defect high confidence hypotheses during inference as ground truth labels. We then conduct training data augmentation experiments using these de-identified datasets and demonstrate that score-based selection can result in a 2\% decrease in semantic error rate and 4\%-7\% decrease in domain classification error rate when compared to the baseline technique of random selection. 
\end{abstract}

\section{Introduction}
 The immense progress in deep learning over the past decade has been, in part, driven by the increasing scale of training data \citep{Chinchilla}, model architectures like transformers \citet{Transformers}, and compute used to train the models. However, the science of surfacing \textit{which} examples to include in training data remains a persistent and applicable question. The recently published Platypus family of models \citep{Platypus} outperformed several SOTA open Large Language Models(LLMs) while being trained on a single GPU for only five hours. The reported success at such a low cost appears primarily due to the quality of their smaller dataset which was curated from large pools of open datasets, which reiterates the significance of curating high quality datasets for model training.
 
 Many studies have been conducted on data selection from large pools of data, but there are challenges when it comes to implementing them in large scale systems. One challenge is that they are studied in an offline, one-time selection setting from a static data pool while  most real world systems need to implement data selection on a continual basis with potential data drift. To make data selection practical and scalable, it should be based on scores that are easy to compute and interpret, stay relevant with changes to data distribution and model architectures, and can be integrated into existing data collection pipelines easily. A second challenge is that some data selection techniques are too compute intensive \citep{BADGE} to be easily implemented and integrated. Third, a few selection methods like "Selection Via Proxy" \citep{SelectViaProxy} require training and maintaining proxy models for data selection, which adds to the resources overhead. In the context of langauge models (LMs), recent studies for data selection experimented with changes to pre-training, such as Task Adaptive Pre-Training (TAPT) \citep{TAPT4AL}. These approaches are not architecture agnostic and the complimentary models need to be retrained with changes to underlying training data, prohibiting continuous data selection without interruptions.

The above mentioned challenges become more pronounced when dealing with unstructured data in low resource languages. Most data selection methods are developed where training data is in English and experiments on non-English languages are typically conducted after the methods are optimized on English data. While this is a gap in research, it also raises the question, "Do these selection methods work for training models on low-resource languages? If so, how well?" There have been studies on data selection methods in multilingual settings for Neural Machine Translation \citep{DynamicDataSelectionNMT}. \citet{hedderich2021survey} surveyed  methods that enable
learning when training data is sparse which includes data augmentation. However, data selection strategies for unstructured data in low-resource languages for supervised machine learning models is relatively a less explored area of research.

All the above challenges and issues emphasize the need to implement efficient and scalable data selection methods for low resource languages. We conduct experiments on Portuguese language. Our work aims to test two metrics -- entropy and EL2N -- in large scale conversational systems. The implementation of these metrics are agnostic to changes to data distribution and model architecture. They are interpretable, easy to compute, scalable, and can be integrated into existing pipelines that involve routine data selection. Our work presents a comprehensive benchmarking of improvements from score based data selection methods, dives deep into how those affect training data, and the overlap of the data selected using different methods for supervised machine learning models.

We conduct our experiments on a BERT based model \citep{BERT} that performs domain, intent and slots recognition. Components that are targeted for improvements are detailed in sec. ~\ref{environment}.

\section{Scoring metrics}
\textbf{Entropy:} Natural Language Understanding (NLU) is a key component in a conversational system. Domain Classifier (DC) within NLU predicts the domain a user request needs to be routed to. A DC can be expressed as a function parametrized by a set of weights $\theta: f_\theta^j(x)$ is the softmax output for $j$ belonging to one of $N$ classes. For some input $x$, we compute entropy \citep{Shannon} of DC outputs for a single example as shown in the eq.~\ref{eqn:entropy}.

\begin{minipage}{.5\linewidth}
\begin{equation}
H(x) = -\sum_{j=0}^{N-1} f_\theta^j(x) \log_2 f_\theta^j(x)
\label{eqn:entropy}
\end{equation}
\end{minipage}%
\begin{minipage}{.5\linewidth}
\begin{equation}
    \texttt{EL2N}(x) = \|\boldsymbol{f}_\theta(x) - \boldsymbol{y}_i\|_2 \, \label{eqn:el2n}
\end{equation}
\end{minipage}

\textbf{EL2N} is a margin-based metric that estimates gradient norms as described in \citet{EL2N} . It is computed as in eq.~\ref{eqn:el2n}, where $\boldsymbol{f}(x)$ indicates the softmax of the model outputs and $\boldsymbol{y}_i$ indicates the one-hot encodings of the label for the $i$th example. Typically, this metric is averaged over $\mathcal{O}(1)$ number of ``replicates" (model initializations) to obtain a reliable signal of example difficulty. We compute EL2N scores at fine-tuning of DC, averaged over five replicates for each example.

Predictions from DC, along with intent and token classification, feed as inputs to downstream tasks in a NLU system. Any improvements to DC could translate to improvement in the entire system. So, we anchored our experiments on data selection with EL2N and entropy based on DC outputs.

\textbf{NLU Model Confidence Score:} In the conversational system under experimentation, NLU Model Confidence Score, referred to as "NLU Score" from here on, is a measure of the system's confidence in the suggested hypothesis i.e, predicted classes (domain and intent) and labels recognition (for individual tokens). NLU Score is a calibrated metric with a range (0,1]. DC scores i.e, softmax outputs from the classifier, are one set of inputs to NLU score. Other inputs include, but are not limited to, intent classifier and token recognition scores. In general, a hypothesis with the correct domain should have a high NLU Score and vice-versa. NLU score is calculated per domain. Unlike softmax outputs from a  classifier, NLU scores from the system across domains don't add to 1.  

\section{Datasets}
\textbf{Existing training data} is the de-identified data used to train an in-house model used for experiments.

\begin{figure}[!htb]
   \begin{minipage}{0.45\textwidth}
     \centering
      \includegraphics[height=0.17\textheight]{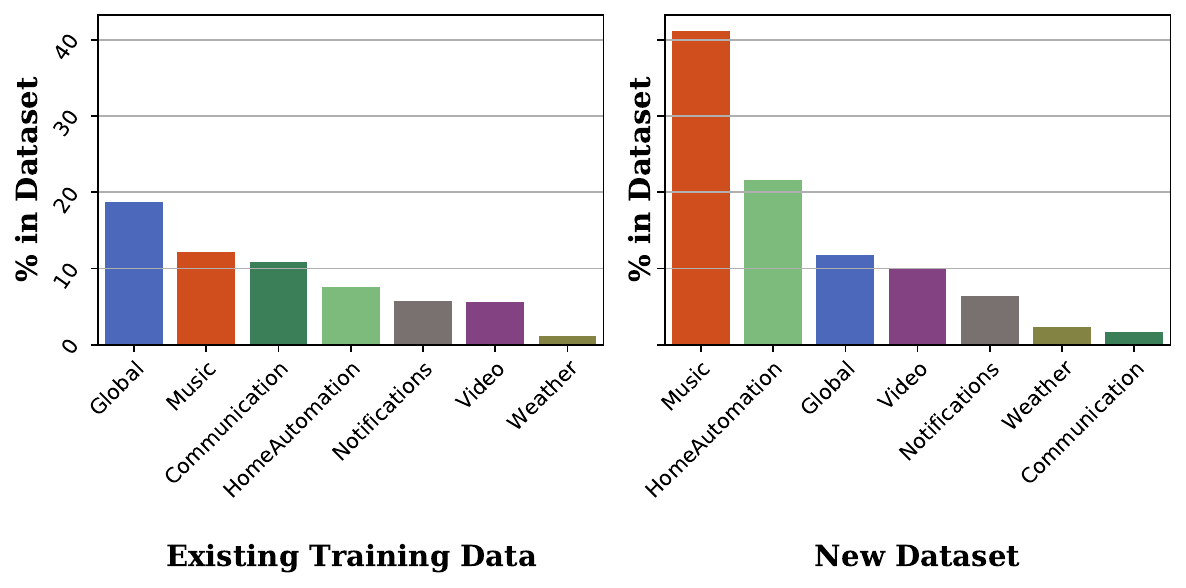}
     \caption[width=0.5\linewidth]{ Data distribution of top seven domains in the existing training data (left) and new dataset (right); Both plots share y-axis.}\label{fig:existing_vs_wsl_data_dist}
   \end{minipage}\hfill
   \begin{minipage}{0.45\textwidth}
     \centering
      \includegraphics[height=0.18\textheight,keepaspectratio]{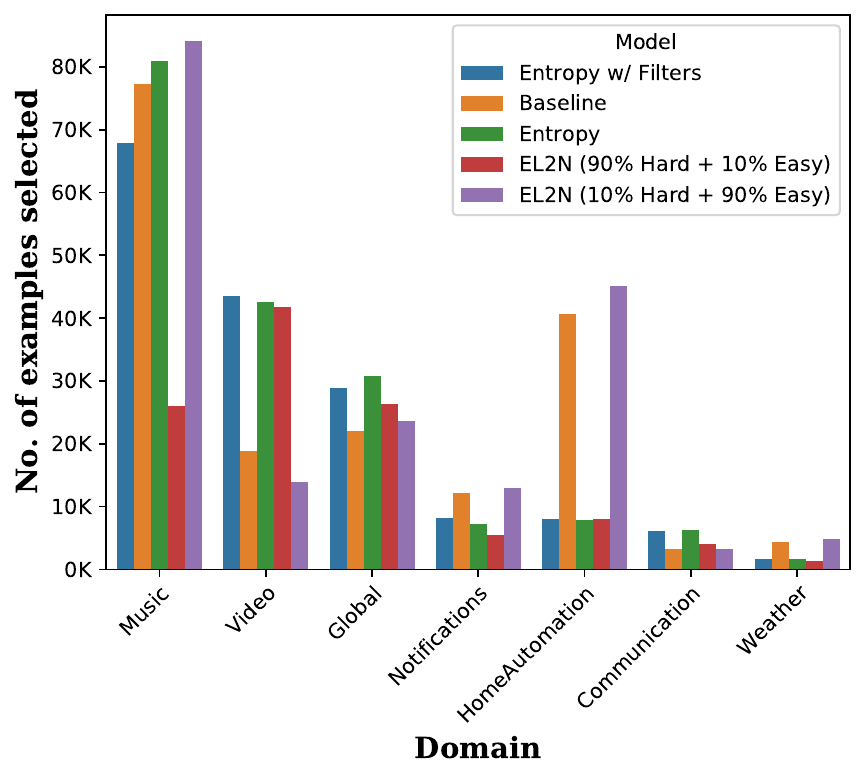}
  \caption[width=0.1\linewidth]{Data distribution of top seven domains in the  datasets curated using different data selection methods}
  \label{fig:dist_of_resultant_datasets}
   \end{minipage}
\end{figure}

\textbf{New dataset}\label{new_data_set} is an inhouse Weak-Signal Labeled (WSL) dataset sourced using the procedure described in \citet{WSL}, where data is constructed from \textit{weak supervision} from the user to obtain NLU labels. For example, if an unsuccessful action from the voice assistant is followed by a user rephrasing their request, which then results in an uninterrupted response from the device, that utterance is pseudolabeled using the top NLU hypothesis. We begin with a de-identified WSL dataset with NLU Score range [0.3,0.85] collected over a period of time, referred to as "new dataset" from here on. The selected range aims to eliminate examples that are already well learnt by the model (NLU score > 0.85) and noisy/ambiguous examples (NLU Score < 0.3). New dataset comes from de-ientified live traffic with its size quite larger when compared to existing training data, and is heavily skewed towards popular user interactions i.e., less diverse. We experimented with entropy and EL2N score based data selection strategies to curate smaller datasets from the new dataset with most useful/difficult examples. These curated datasets are added to the existing training data to build different candidate models as listed in sec. ~\ref{candidate_models}. Figure ~\ref{fig:existing_vs_wsl_data_dist} shows data distribution of top seven domains, accounting to 90\% user traffic, in existing training data (left) and new dataset (right). Figure ~\ref{fig:dist_of_resultant_datasets} shows the distribution of datasets curated using score based selection methods detailed in sec.~\ref{candidate_models}.

\textbf{Dataset curation based on Entropy:} \label{entropy_dataset_curation}
For each sample in the new dataset we generate DC scores for possible domains using the eq. ~\ref{eqn:entropy}, and calculate entropy of the generated DC scores. We then rank all the examples based on their entropy scores and select top \textit{K} examples to create a smaller dataset. Sec.~\ref{entropy_selection_details} has exact details of how the examples are selected based on entropy.

\textbf{Dataset curation based on EL2N:} \label{el2n_dataset_curation}
For each training example in the dataset, we generate EL2N scores using eq. ~\ref{eqn:el2n} averaged over five replicates. The scores are generated through a domain classification task. After the data was ranked by EL2N difficulty, we use a threshold score of $\le 0.15$ to denote ``easy" examples and a threshold of $\ge 0.6$ to denote ``hard" examples. We then randomly sampled from these subsets for a given target dataset size to obtain different mixtures of easy and hard examples, with splits described in sec. \ref{candidate_models}.

\textbf{Specialized testsets:}
\label{specialized_testsets}
We use specialized test sets to evaluate model performance for the use cases of interest. Our objective is to improve generalizability of the model, and so we evaluate our models on specially curated test sets that have the same distribution as the tail 40\% of user traffic and are de-identified.

\section{Experiments}
\subsection{Models}
\label{candidate_models}
We take one of the in-house models that performs NLU tasks as a baseline to experiment for further accuracy improvements. The in-house model is retrained  with different additional datasets, all of the same size, resulting in different candidate models as listed below.

\textbf{Baseline model} has existing training data augmented with randomly selected data from new dataset. 

\textbf{EL2N (10\% Hard + 90\% Easy)} candidate has training data augmented with data selected based on EL2N scores from the new dataset with a composition of 10\% hard examples and 90\% easy examples. Section \ref{el2n_dataset_curation} has details on how easy and hard examples are selected.

\textbf{EL2N (90\% Hard + 10\% Easy)} candidate has training data augmented with data selected based on EL2N scores from the new dataset with a composition of 90\% hard examples and 10\% easy examples. Section \ref{el2n_dataset_curation} has details on how easy and hard examples are selected.

\textbf{Entropy} candidate has training data augmented with data selected based on entropy scores of domain classifier outputs as detailed in sec.~\ref{entropy_dataset_curation}. Examples with higher entropy scores are added to  training data. Sec.~\ref{entropy_selection_details} has exact details of why and how the examples were selected based on entropy.

\textbf{Entropy with Filters} candidate has additional data processing on the dataset curated based on entropy scores. First, we limit the repetition of an example in the curated dataset to \textit{P} to enhance the diversity. We observed that with no limit in place, approx. 4,000 examples selected for a domain came from two unique examples. Second, we ensure that each domain has a minimum representation of \textit{R}\% in the dataset. Our values for \textit{P} and \textit{R} are 20 and 0.5. Further discussion can be found at sec. ~\ref{entropy_selection_details}.

\subsection{Evaluation Metrics}

We measure candidates' performance on specialized test sets introduced in sec. \ref{specialized_testsets} in terms of component-wise (domain, intent and slots) error rates. In terms of component-wise error rates, we measured domain classification performance using the recall-based classification error rate \texttt{DCER}. To evaluate slot-filling performance, we measured semantic error rate (\texttt{SEMER\label{semer_eq}}):
\[
\small
\texttt{SEMER} \equiv \dfrac{\texttt{\# Intent errors} + \texttt{\# Slot errors}}{\texttt{\# Test data} + \texttt{\# Slots}}.
\]
We measured F-SEMER, the harmonic mean of SEMER using predicted labels as the reference and SEMER computed on ground-truth labels as the reference; this score balances precision/recall equally. We also report the interpretation error rate \texttt{IRER}, which reflects the rate of any kind of error (slots, intents, domain).

\section{Results}
 Table ~\ref{tab:overall_evaluation_results} presents evaluation results for different candidates listed in the section \ref{candidate_models}. All the values are relative to a
baseline model trained on randomly sampled data from the new dataset. We see improvement across all metrics for all candidates except \textit{EL2N (90\% Hard + 10\% Easy)}. Improvements to SEMER, F-SEMER and IRER are in the order of 2\% with respect to the baseline. Improvements to DCER are in the range of 3\%-7\%. \textit{Entropy w/ Filters} candidate and \textit{EL2N (10\% Hard + 90\% Easy)} have the most promising results. 

\begin{table*}[h]

\caption{Evaluation results relative to a baseline with random data selection  \label{tab:overall_evaluation_results}}

\centering
\begin{tabularx}{\textwidth}{ccccc}
\toprule
Model & $\Delta$ SEMER\% $\downarrow$ & $\Delta$ F-SEMER\% $\downarrow$ & $\Delta$ DCER\% $\downarrow$ & $\Delta$ IRER $\downarrow$ \\
\midrule

Entropy & -2.35 & \textbf{-2.29} & -6.11 & -1.46 \\
Entropy w/ Filters & \textbf{-2.37} & -2.24 & \textbf{-7.20} & -0.55 \\
EL2N (90\% Hard + 10\% Easy) & -0.08 & -0.16 & 3.09 & -0.06 \\
EL2N (10\% Hard + 90\% Easy) & -2.14 & -2.16 & -4.12 & \textbf{-2.10} \\
\bottomrule
\end{tabularx}
 
\end{table*} 

Table \ref{tab:domain_level_evaluation_results} presents more nuanced results when we look at domain level metrics. One can notice that Video, Notifications, Weather and Communications domains are better served by entropy based data selection while Music and Home Automation domains are better served by EL2N score based data selection. We recommend avoiding "one-size-fits-all" approach and encourage identifying which technique performed the best for each domain. Once the initial experiments are done, domains can be mapped to different data selection pipelines. Sec.~\ref{metrics_by_domain_section} and sec.~\ref{appdx_analysis} have further discussion on metrics and data selected.

\section{Conclusion}

Entropy based data selection improved metrics (DCER) better than EL2N on anchor task (DC), which was the source for the score. EL2N, however, delivered better improvements to overall recognition as measured by IRER. We hypothesize that EL2N, averaged over multiple runs, captures example importance towards overall accuracy, while entropy is better suited to improve a specific task. With growing developments in LLMs, we plan to continue our work on data selection strategies for LLMs, and using LLMs for data selection. Our areas of research would be fine-tuning LLMs, in-context learning \citep{brown2020language}, avoiding model collapse \citep{ModelCollapse}, and experimenting with a broader set of metrics (\citet{EL2NinLLMs}, \citep{DiversityCoefficient}) for data selection.



\bibliography{references.bib}

\appendix

\section{Environment: \label{environment}} We conducted our experiments on a conversational system. A typical conversational system has speech and Natural Language Understanding (NLU) components along with many others such as business logic based hypothesis re-routing, invoking third party applications, and more. Our experiments target improvements to NLU components. Within NLU, we need to recognize multiple things correctly to deliver desired experience to end users. A few of them are \textit{domain, intent, and slots}. In the example, "Play Taylor Swift", \textit{domain} is "Music", \textit{intent} is "Play Music" and \textit{slot\_name} is "Artist Name" with a \textit{slot\_value} of "Taylor Swift". In our results, we show how our techniques improved metrics on each of these recognition tasks. Our experiments are anchored on domain classifier outputs. 

\section{Exploratory Data Analysis}

\subsection{Correlation study of NLU scores and Entropy scores}
We did a correlation study between NLU scores and corresponding entropy of domain classifier scores on a pool of 5 million examples. Results are shown in Table \ref{tab:correlation}.

All the three correlation coefficients show a consistent negative correlation between model confidence scores and entropy scores which implies that by reducing entropy of DC scores, we could improve NLU models' confidence on the correct interpretations and serve the end users with the most relevant responses.

\begin{table}[ht!]
\centering
\caption{\label{tab:correlation}Correlation coefficients between entropy of domain classifier scores and NLU scores}
\begin{tabular}{lc}
\toprule
Correlation Coefficient & Value\\
\midrule
Pearson Correlation & -0.3132\\
Spearman’s Rank Correlation & -0.0732\\
Kendall tau-a & -0.104\\
\bottomrule
\end{tabular}
\end{table}

\subsection{Distribution of Entropy and EL2N on the large new dataset}

\begin{figure}[!htb]
   \begin{minipage}{0.45\textwidth}
     \centering
      \includegraphics[width=1.25\linewidth,keepaspectratio]{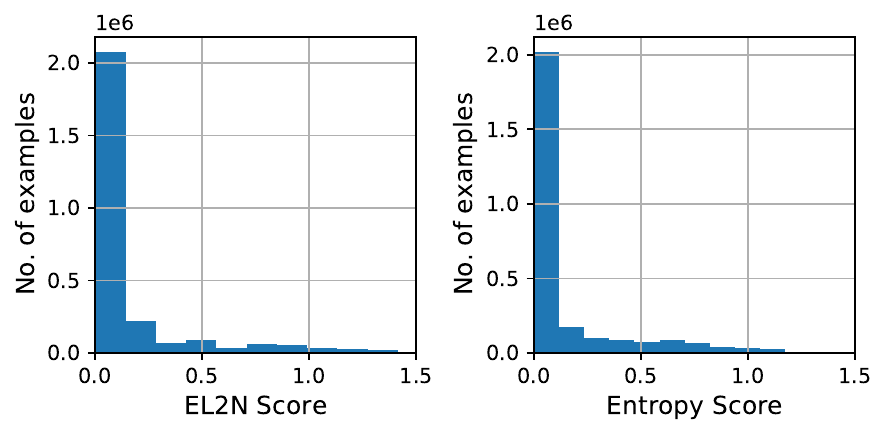}
     \caption[width=0.5\linewidth]{EL2N and Entropy scores distribution on the new dataset.}\label{fig:Entropy_n_El2n_dist}
   \end{minipage}\hfill
   \begin{minipage}{0.45\textwidth}
     \centering
      \includegraphics[width=0.75\linewidth,keepaspectratio]{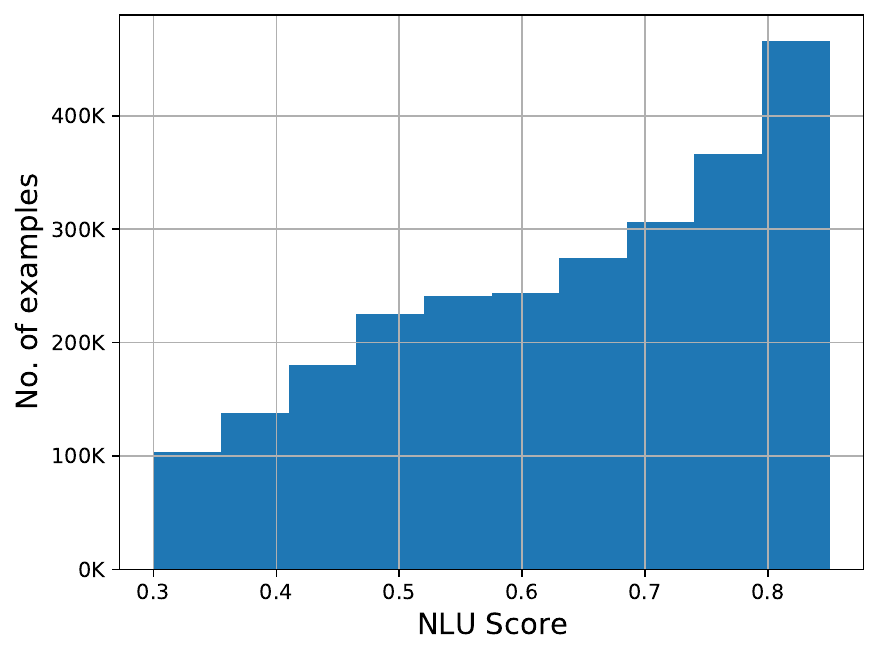}
  \caption[width=0.5\linewidth]{NLU Score distribution in the new dataset}
  \label{fig:nlu_score_dist}
   \end{minipage}
\end{figure}

Both Entropy and EL2N metrics have long tail distributions on the new dataset (Figure ~\ref{fig:Entropy_n_El2n_dist}). Most of the examples have low scores implying relative high certainty in prediction (low entropy) or relative ease of getting correct prediction (low EL2N). We can observe that Entropy and EL2N scores distributions are left skewed in the Figure ~\ref{fig:Entropy_n_El2n_dist} while NLU score distribution in the Figure ~\ref{fig:nlu_score_dist} is right skewed re-iterating the they trend in the opposite directions. Hence, reducing entropy or EL2N scores on a dataset could improve NLU score and eventually the system. The long tail presents a real opportunity to select and add challenging examples to training data. Models benefit from learning from these challenging examples during training which could result in improved accuracy. 

Range of Entropy for the full dataset after removing outliers (z-score > 3) is [0.0, 1.1731] with a mean and standard deviation of 0.137 and 0.252 respectively. Range of EL2N score for the full dataset after removing outliers (z-score > 3) is [0.0, 1.414] with a mean and standard deviation of 0.128341 and 0.275945 respectively.

\section{Scoring metrics \label{appndx_scoring_metrics}}

\textbf{Entropy} $H$ of a random variable $X$ is the level of uncertainty inherent in the variable’s possible outcomes. In case of a classification problem, $p_j(x_{i})$ is the probability of an example $x_{i}$ belonging to the class \textit{j}. The entropy of a single example $x_{i}$ that could belong to \textit{n} possible classes is calculated as \begin{equation}
    H(X=x_i) = -\sum_{j=1}^{n} p_j(x_i) \log_2 p_j(x_i)
\end{equation}

\subsection{Motivation to use Entropy for data selection}
In Active Learning (AL), entropy is used as an acquisition function to find the most \textit{informative} examples to label from a large pool of unlabelled data. Entropy captures the uncertainty associated with the predicted labels. The higher the entropy of predictions, the higher is the possibility that the model is \textit{more challenged or confused} when presented with such samples. Samples with higher entropy are selected for annotations. Those annotated examples are then added to training data to improve the models. In AL, Entropy based selection is proved to work better than many other sophisticated acquisition functions like BADGE (\citet{BADGE}) and BALD (\citet{BALD}) when tested on a variety of tasks. Recent methods like AcTune (\citet{AcTune}) show that entropy when employed with other scientific techniques delivers superior results. Drawing inspiration from entropy based data selection for labeling in AL, we implemented entropy as a metric to select data from a large pool of \textit{Weak-Signal Labeled(WSL)} data. More information on WSL data is presented in section \ref{new_data_set} . One more reason to choose entropy over other methods is the ease of implementation and cost to compute.

\textbf{Selection based on entropy: \label{entropy_selection_details} } For any new dataset \textit{D}, we compute entropy of domain classification softmax outputs, referred to as "entropy". We rank examples in \textit{D} in decreasing order of their entropy and select top \textit{K} examples based on a given cut off criteria. Intuition behind this is that we train the models with data that the models are less certain about but would benefit from learning from those during training. This should result in improved accuracy of the model predictions. Our cut off criteria was to limit the number of additional training examples that can be added to the existing training data. We experimented with newly created datasets to be in the order of 2\% or 5\% of existing training data. Datasets of size 5\% gave the best results.  Other selection criteria could be cut off based on absolute entropy score, top X\% of examples from the larger dataset, etc.\\

For \textbf{Entropy w/ Filters} candidate, we chose values of 20 and 0.5\% for an example repetition cap and minimum representation of a domain (class). We arrived at these values after multiple experiments. There values can and should be experimented for different modeling tasks and datasets.

\section{Results}

\subsection{Metrics by Domain \label{metrics_by_domain_section}}
Table ~\ref{tab:domain_level_evaluation_results} presents evaluation results for the top seven domains (classes) that account to 90\% user traffic. Results show that different domains benefit from different data selection strategies. We recommend avoiding "one-size-fits-all" approach and encourage identifying which technique performed the best for each domain. Once the initial experiments are done, domains can be mapped to different data selection pipelines. Looking at the Table ~\ref{tab:domain_level_evaluation_results}, one can notice that Video, Notifications, Weather and Communications domains are better served by entropy based data selection while Music and Home Automation domains are better served by EL2N score based data selection. For domain level metrics, we reported balanced metrics for DCER and IRER i.e, F-DCER and F-IRER along with SEMER that was introduced in sec.~\ref{semer_eq}

\begin{table*}[ht]
 \caption{Evaluation results by domain relative to a baseline with random data selection}
  \label{tab:domain_level_evaluation_results}
\begin{tabular}{ccccc}
\toprule
Domain & Model & $\Delta$ SEMER \%  & $\Delta$ F-DCER\% & $\Delta$ F-IRER \\
\midrule
\multirow{4}{6em}{Music} & Entropy & -5.15 & -5.38 & -1.92 \\
& Entropy w/ Filters & -3.34 & -5.83 & -0.41 \\
& EL2N (90\% Hard + 10\% Easy) & -1.80 & 7.17 & -1.11 \\
& EL2N (10\% Hard + 90\% Easy) & \textbf{-4.03} & \textbf{-5.83} & \textbf{-2.85} \\
\midrule
\multirow{4}{6em}{Video} & Entropy & 1.26 & -10.03 & -3.99 \\
& Entropy w/ Filters & \textbf{-1.73} & \textbf{-11.34} & \textbf{-3.99} \\
& EL2N (90\% Hard + 10\% Easy) & 1.73 & 1.97 & 1.56 \\
& EL2N (10\% Hard + 90\% Easy) & 3.06 & -4.98 & -1.82 \\
\midrule
\multirow{4}{6em}{Home Automation} & Entropy & -0.93 & -4.07 & 1.43 \\
& Entropy w/ Filters & -2.51 & -6.50 & 2.90 \\
& EL2N (90\% Hard + 10\% Easy) & 1.02 & 0.81 & 1.52 \\
& EL2N (10\% Hard + 90\% Easy) & \textbf{-3.62} & \textbf{-5.28} & \textbf{-2.43} \\
\midrule
\multirow{4}{6em}{Global} & Entropy & -4.82 & -0.12 & 2.14 \\
& Entropy w/ Filters & 1.08 & -1.09 & 0.78 \\
& EL2N (90\% Hard + 10\% Easy) & 1.58 & 7.13 & 3.18 \\
& EL2N (10\% Hard + 90\% Easy) & 2.81 & -2.06 & -0.71 \\
\midrule
\multirow{4}{6em}{Notifications} & Entropy & -4.18 & 1.45 & 2.79 \\
& Entropy w/ Filters & \textbf{-6.07} & \textbf{-0.72} & \textbf{2.79} \\
& EL2N (90\% Hard + 10\% Easy) & 2.72 & 25.36 & 2.15 \\
& EL2N (10\% Hard + 90\% Easy) & -5.23 & -0.72 & 0.75 \\
\midrule
\multirow{4}{6em}{Weather} & Entropy & \textbf{-3.50} & \textbf{-10.46} & \textbf{0.69} \\
& Entropy w/ Filters & -3.15 & -4.58 & 5.02 \\
& EL2N (90\% Hard + 10\% Easy) & 3.85 & 1.96 & 7.49 \\
& EL2N (10\% Hard + 90\% Easy) & 2.45 & -3.92 & 3.15 \\
\midrule
\multirow{4}{8em}{Communications} & Entropy & -0.50 & 2.27 & 1.07 \\
& Entropy w/ Filters & \textbf{-0.83} & \textbf{-1.55} & \textbf{-0.09} \\
& EL2N (90\% Hard + 10\% Easy) & 1.87 & 1.09 & 1.86 \\
& EL2N (10\% Hard + 90\% Easy) & 0.96 & -1.18 & 0.42 \\
\bottomrule
\end{tabular}
\end{table*}

\section{Analysis \label{appdx_analysis}}
\subsection{Data Distribution of Selected Training Examples by Each Method}
Figure \ref{fig:dist_of_resultant_datasets} shows the data distribution of top seven domains, accounting to 90\% user traffic, in the datasets curated based on different data selection methods. Baseline method randomly selects examples from the new dataset and is representative of user traffic. Datasets curated for \textit{Entropy} and \textit{Entropy w/ Filters} have similar distribution across domains. This is corroborated by entropy datasets' overlap of 89.9\% presented in Table \ref{tab:datasets_overlap}. These two methods tend to surface diverse examples on which the models are less certain about the predictions. As a result, their distribution could vary from baseline's. We observe that entropy based methods picked less no. of examples for the domain \textit{HomeAutomation} relative to baseline as the user interactions are relatively more standard i.e., less diverse. An example interaction for \textit{HomeAutomation} is "Turn of the lights". They picked more no. of examples for \textit{Video, Global} and \textit{Communication} given the diverse nature of user interactions. Example interactions for \textit{Video} are "Play my favorite movie" and "I want to watch Harry Porter". Datasets from EL2N based selection don't have similar distributions unlike entropy based selection. This is corroborated by EL2N datasets' overlap of 19.7\% presented in Table \ref{tab:datasets_overlap}. It is interesting to note that EL2N (10\% Hard + 90\% Easy) selected more examples in six out of seven domains under review relative to baseline while EL2N (90\% Hard +10\% Easy) has shown no clear pattern.

\subsection{Overlap of data between Entropy and EL2N score based selections}
Table \ref{tab:datasets_overlap} presents the overlap of data between datasets curated by different methods. Entropy based datasets had the highest overlap of 89.94\%. This is not surprising given they both select examples based on their ranked entropy scores and there is no randomness involved. They only differ for limits of example repetition and minimum domain representation. On the other hand, EL2N based datasets have significantly lower overlap of 19.7\% when compared to entropy datasets. This is because these datasets are randomly sampled from different regions of EL2N score distribution. Except for \textit{EL2N (90\% Hard + 10\% Easy)}, all candidates delivered superior results when compared to the baseline. We hypothesize that this is because the said method deliberately adds "difficult" examples in large volume (90\%) which could result in adding more noise than anticipated, and making the model convergence slower within give no. of training epochs.

\begin{table*}[ht]
\caption{Overlap (percent) of data between datasets curated by different methods}
 \label{tab:datasets_overlap}
 \resizebox{\textwidth}{!}{%
\begin{tabular}{cccccc}
\toprule
& Baseline & Entropy w/ Filters & Entropy & EL2N (10\% Hard + 90\% Easy) & EL2N (90\% Hard + 10\% Easy)\\
\midrule
Baseline & 1 & 7.4 & 7.7 & 7.6 & 7.2 \\ \hline
Entropy w/ Filters & & 1 & 89.94 & 6.7 & 36.6 \\ \hline
Entropy & & & 1 & 7.1 & 33.57 \\ \hline
EL2N (10\% Hard + 90\% Easy) & & & & 1 & 19.7 \\ \hline
EL2N (90\% Hard + 10\% Easy) & & & & & 1 \\ 
\bottomrule
\end{tabular}}

\end{table*}

\section{Limitations}
When opting for entropy based data selection one needs to understand that (i). some level of uncertainty or diversity in responses is inherent in conversational systems and is desired. Response's success is subject to user's preference, location and context. For example, "Resume Harry Porter" could be interpreted as "resume the Harry Porter movie I was watching on my TV earlier" if the user has an active video session in the environment he is interacting with. If not, it could be interpreted as "resume playing an audio book on Harry Porter" if the user is interacting with a screen less device. So, augmenting training data with one particular interpretation where multiple correct responses exist is something practitioners need to watch out for. It is recommended that we augment data with all possible or most relevant interpretations for the same input, (ii) increasing NLU confidence score might not translate to correct prediction in all cases. If we are not watchful, we could be adding noise i.e, examples that are ambiguous or corrupt. This could potentially degrade models' performance. One way to address this problem is to exclude outliers based on entropy score (z-score >= 3). In addition to excluding outliers, we can instate a criteria of minimum repetition in the dataset to avoid selecting such one off unusual user interactions.

When it comes to EL2N based data selection, one needs to experiment with different thresholds and data compositions before deciding on the optimal values. These optimal values could change with model architecture and data drift.

\end{document}